\documentclass[manuscript]{acmart}
\AtBeginDocument{%
  }

\setcopyright{acmlicensed}
\copyrightyear{2018}
\acmYear{2018}
\acmDOI{XXXXXXX.XXXXXXX}
\acmConference[Conference acronym 'XX]{Make sure to enter the correct
  conference title from your rights confirmation email}{June 03--05,
  2018}{Woodstock, NY}
\acmISBN{978-1-4503-XXXX-X/2018/06}

\begin{document}


\title[Assuring CPS with Foundation Models]{Assuring Learning-Enabled Cyber-Physical Systems with Foundation Models: A Neuro-Symbolic Agenda for Testing, Verification, and Runtime Assurance}


\author{Xi Zheng}
\affiliation{%
  \institution{Macquarie University}
  \city{Sydney}
  \country{Australia}}
\email{xi.zheng@mq.edu.au}

\author{Ivan Ruchkin}
\affiliation{%
  \institution{University of Florida}
  \city{Gainesville}
  \country{USA}}
\email{iruchkin@ece.ufl.edu}

\author{Hong Jin Kang}
\affiliation{%
  \institution{University of Sydney}
  \city{Sydney}
  \country{Australia}}
\email{hongjin.kang@sydney.edu.au}

\author{Tevfik Bultan}
\affiliation{%
  \institution{University of California, Santa Barbara}
  \city{Santa Barbara}
  \country{USA}}
\email{bultan@ucsb.edu}

\author{Vijay Ganesh}
\affiliation{%
  \institution{Georgia Institute of Technology}
  \city{Atlanta}
  \country{USA}}
\email{vganesh45@gatech.edu}

\author{Taylor T. Johnson}
\affiliation{%
  \institution{Vanderbilt University}
  \city{Nashville}
  \country{USA}}
\email{taylor.johnson@vanderbilt.edu}

\author{Ziyang Li}
\affiliation{%
  \institution{Johns Hopkins University}
  \city{Baltimore}
  \country{USA}}
\email{ziyang@cs.jhu.edu}

\author{Lei Ma}
\affiliation{%
  \institution{The University of Tokyo}
  \city{Tokyo}
  \country{Japan}}
\email{malei@is.s.u-tokyo.ac.jp}

\author{Aloysius K. Mok}
\affiliation{%
  \institution{The University of Texas at Austin}
  \city{Austin}
  \country{USA}}
\email{mok@cs.utexas.edu}

\author{Daniel Neider}
\affiliation{%
  \institution{TU Dortmund University}
  \city{Dortmund}
  \country{Germany}}
\email{daniel.neider@tu-dortmund.de}

\author{Ruzica Piskac}
\affiliation{%
  \institution{Yale University}
  \city{New Haven}
  \country{USA}}
\email{ruzica.piskac@yale.edu}

\author{Nalini Venkatasubramanian}
\affiliation{%
  \institution{University of California, Irvine}
  \city{Irvine}
  \country{USA}}
\email{nalini@uci.edu}

\author{Simon Thompson}
\affiliation{%
  \institution{TIER IV}
  \city{Tokyo}
  \country{Japan}}
\email{simon.thompson@tier4.jp}

\author{Sanjoy Baruah}
\affiliation{%
  \institution{Washington University in St. Louis}
  \city{St. Louis}
  \country{USA}}
\email{baruah@wustl.edu}

\renewcommand{\shortauthors}{Zheng et al.}

\newcommand{\JZ}[1]{\par\smallskip\noindent{\color{blue}[James: #1]}\par\smallskip}

\begin{abstract}
Learning-enabled cyber-physical systems (CPS), including autonomous vehicles and UAVs, increasingly incorporate foundation models, yet fragmented assurance practices provide insufficient support for system-level guarantees. We argue that restoring engineering control across stochastic neural, symbolic, and physical boundaries requires a shift from model-centric evaluation to system-level assurance. We introduce the Foundation Model Assurance Stack, connecting three directions: (i) \emph{neuro-symbolic architectures} that constrain foundation models through explicit symbolic world models, reasoning layers, and interfaces; (ii) \emph{semantically grounded, conformance-based testing} that combines formal specifications and symbolic world models with foundation-model-assisted specification mining, scenario generation, and test synthesis; and (iii) \emph{compositional verification and runtime assurance} that target formally tractable components and interfaces through modular contracts, equivalence checking, verified synthesis, uncertainty-aware reasoning, and runtime monitoring and enforcement. Symbolic abstractions establish assurance boundaries between stochastic learning and formal reasoning, connecting testing, verification, and runtime evidence across the system lifecycle. Foundation models serve as constrained assistants whose outputs are independently checked wherever correctness can be formally specified. We identify research challenges in specification engineering, abstraction fidelity, evidence composition, and lifecycle certification, and outline a staged agenda toward trustworthy, deployable, and certifiable learning-enabled CPS.

\end{abstract}



\received{xxxx}
\received[revised]{xxxx}
\received[accepted]{xxxx}

\maketitle

\section{Introduction}
Learning-enabled cyber-physical systems (CPS), including autonomous vehicles,
unmanned aerial vehicles (UAVs), and embodied robotic systems, increasingly
integrate foundation models such as large language models (LLMs),
vision-language models (VLMs), and vision-language-action models (VLAs).
These models provide powerful capabilities for perception, semantic
understanding, reasoning, planning, and system engineering. At the same time,
their stochastic, opaque, and data-driven behaviour challenges fundamental
software engineering assumptions about explicit specifications, predictable
interfaces, traceable execution, and independently checkable correctness.


The resulting assurance problem extends beyond the reliability of individual
models. In a learning-enabled CPS, information flows from physical sensors
through neural perception and learned representations to world models,
reasoning and planning, and ultimately back to physical actions. Uncertainty,
semantic information, and behavioural assumptions may be lost or transformed
at each boundary. Consequently, strong benchmark performance of an individual
model does not establish that the integrated system behaves correctly, remains
within its operational constraints, or fails safely under distribution shift
and unforeseen environmental conditions. Even formally verified model
properties provide practical assurance only when they are connected to the
requirements of the systems in which those models are
deployed~\cite{johnson2025allerton}. Establishing this connection requires
relating model-level guarantees to formal system specifications and
environmental assumptions, consistent with the broader verified-AI
agenda~\cite{seshia2022verified}. The central challenge is therefore
one of regaining \emph{engineering control} over learning-enabled systems whose
behaviour cannot be completely specified or analysed at the model level.

This challenge motivates a fundamental question for software engineering:

\begin{quote}
\emph{How can we systematically test, verify, and assure learning-enabled
cyber-physical systems that incorporate foundation models?}
\end{quote}

Discussions at Shonan Seminar No.~235
\cite{zheng2026shonan235} on \emph{LLM-Guided Synthesis, Testing, and
Verification of Learning-Enabled Cyber-Physical Systems} highlighted several
limitations of existing assurance approaches. Testing based primarily on model
accuracy, input diversity, or neural coverage provides limited evidence about
system-level safety properties. Formal verification, meanwhile, faces severe
scalability and specification challenges when applied directly to large
foundation models and open-ended neural components. Treating such models as
black boxes further obscures the semantic relationships between perception,
reasoning, and physical behaviour. A more tractable direction is to expose
explicit intermediate representations and interfaces, use foundation models as
constrained assistants rather than trusted oracles, and concentrate formal
reasoning on components and properties for which correctness can be specified
and independently checked.

Therefore, we argue for a shift from \emph{model-centric evaluation} toward
\emph{system-level assurance}. Rather than attempting to make the foundation
model itself the trusted core of a CPS, software architecture should establish
explicit assurance boundaries around learning-enabled components. In
particular, neuro-symbolic decomposition and symbolic world models provide a
bridge between stochastic neural perception and formally analysable reasoning,
while specifications, contracts, and interfaces provide shared artefacts for
testing, verification, and runtime monitoring. Assurance can then be built
compositionally by combining empirical evidence, formal guarantees, and
runtime enforcement.

\begin{figure*}[t]
    \centering
    \includegraphics[width=0.95\textwidth]{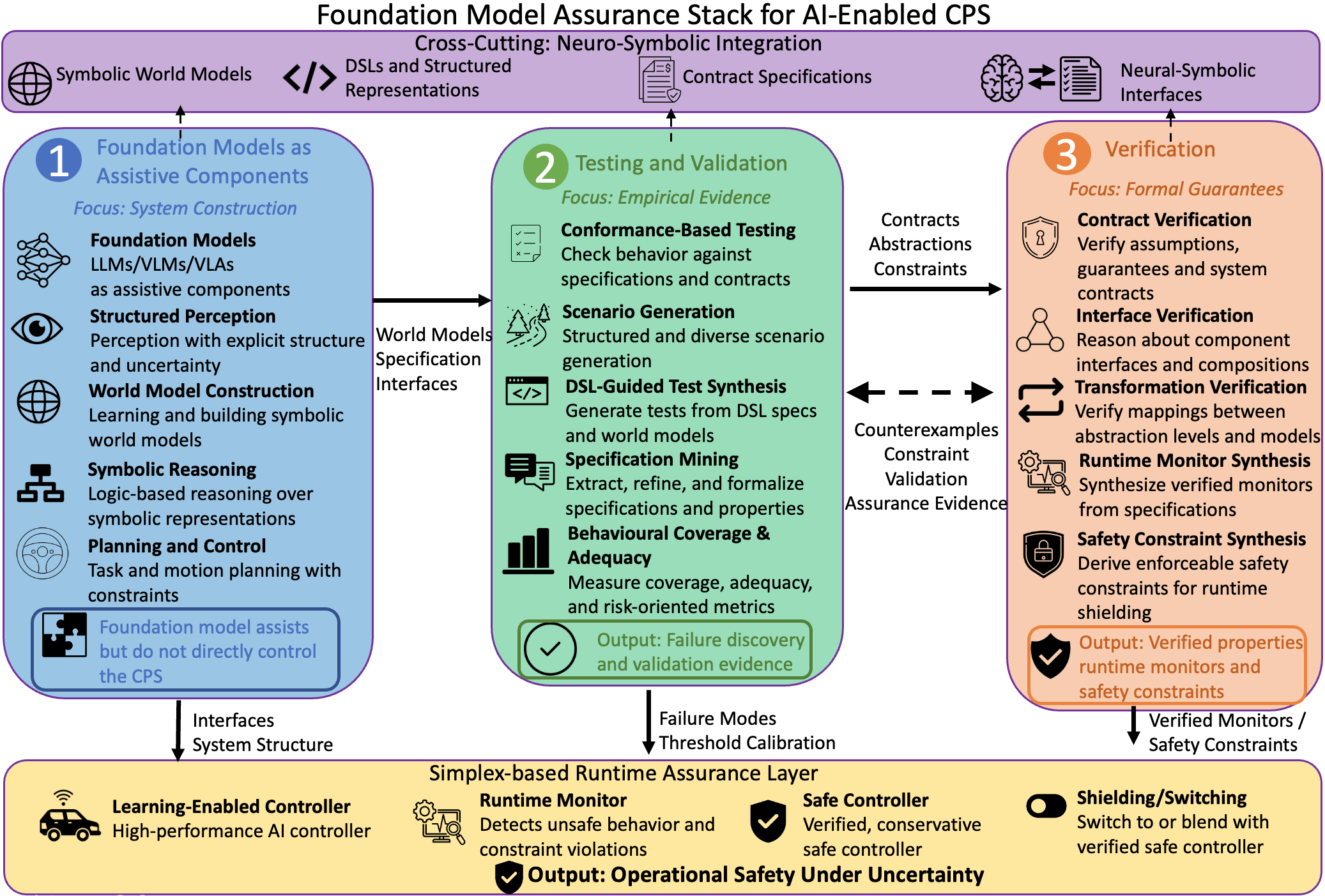}
\caption{\textbf{Foundation Model Assurance Stack for AI-Enabled CPS.}
The stack organizes assurance into three complementary pillars: foundation models as assistive components in CPS architectures, testing and validation of AI-enabled systems, and verification and runtime assurance. These pillars are connected by cross-cutting neuro-symbolic integration, including symbolic world models, DSLs and structured representations, contract specifications, and neural--symbolic interfaces. Architecture provides world models, specifications, and interfaces to testing, and contracts, abstractions, and constraints to verification. Testing provides empirical evidence through conformance-based testing, structured scenario generation, DSL-guided test synthesis, LLM-assisted specification mining, and adequacy assessment. Verification and runtime assurance provide formal guarantees through contract verification, interface verification, transformation verification, runtime monitoring, and safety enforcement. At runtime, a Simplex-style resilience layer combines a learning-enabled controller, runtime monitor, safe controller, and shielding or switching mechanism to support operational safety under uncertainty.}
    \label{fig:assurance-stack}
\end{figure*}

Based on this perspective, we introduce the \emph{Foundation Model Assurance
Stack} shown in Figure~\ref{fig:assurance-stack}. The stack consists of three
complementary assurance pillars, connected through cross-cutting neuro-symbolic
integration, together with a distinct runtime assurance layer that operationalises
their outputs during system execution:

\begin{itemize}

\item \textbf{Foundation Models as Assistive Components} focuses on
\emph{system construction}. Foundation models support structured perception,
symbolic world-model construction, symbolic reasoning, and planning and
control, but do not directly control the CPS. The resulting world models,
specifications, interfaces, and system structure expose explicit artefacts for
downstream assurance.

\item \textbf{Testing and Validation} focuses on \emph{empirical evidence}.
Conformance-based testing, structured scenario generation, DSL-guided test
synthesis, specification mining, and behavioural coverage and adequacy analysis
discover failures and generate validation evidence. Testing also provides
counterexamples and constraint-validation results to verification.

\item \textbf{Verification and Assurance Mechanisms} focuses on
\emph{formal guarantees and assurance artefact synthesis}. Contract
verification, interface verification, and transformation verification establish
properties over tractable system abstractions and interfaces, while runtime
monitor synthesis and safety constraint synthesis translate verified properties
into monitors and enforceable constraints for operational use by the runtime
assurance layer. Verification and testing interact bidirectionally: formal
constraints guide testing, while empirical counterexamples refine verification
artefacts.

\end{itemize}

Across these pillars, \emph{neuro-symbolic integration} provides shared
assurance artefacts through symbolic world models, DSLs and structured
representations, contract specifications, and neural--symbolic interfaces.
These artefacts bridge stochastic learning components and explicit reasoning
and provide common objects for system construction, testing, and verification.

Distinct from the three assurance pillars, the \emph{Runtime Assurance Layer}
operationalises their outputs during system execution through a
Simplex architecture~\cite{Sha2001}. It integrates system structure and
interfaces from the architectural pillar, failure modes and threshold
calibration from testing and validation, and monitors and safety constraints
from verification and runtime assurance. A high-performance learning-enabled
controller operates under a runtime monitor, while a verified safe controller
and shielding or switching mechanisms intervene when unsafe behaviour or
constraint violations are detected. Thus, Pillar~3 develops the verified
properties and enforcement mechanisms needed for runtime assurance, whereas
the bottom layer instantiates and executes these mechanisms to maintain
\emph{operational safety under uncertainty}.

\section{Foundation Models as Assistive Components in CPS Architectures}
\label{secFoundationModels}
This section develops the architectural pillar of the Foundation Model Assurance Stack. We describe how foundation models can support learning-enabled CPS as constrained assistive components, while symbolic world models, explicit neural–symbolic interfaces, and structured reasoning and planning provide the architectural foundations for subsequent testing, verification, and runtime assurance.

\subsection{Structured Perception and World Model Construction}
Foundation models have significantly advanced perception capabilities for learning-enabled cyber-physical systems (CPS), enabling robust object detection, semantic segmentation, scene understanding, and multimodal interpretation from heterogeneous sensor data \cite{firoozi2025foundation}. However, assurance ultimately concerns the correctness of system behaviour rather than the accuracy of neural perception itself. Safety requirements such as collision avoidance, rule compliance, and mission completion are expressed over objects, spatial relationships, temporal evolution, and environmental constraints, rather than pixels or latent neural features \cite{zheng2025neurostrata}. Consequently, the first architectural step in trustworthy CPS is to transform stochastic perception outputs into explicit structured representations that subsequent software components can interpret, reason about, test, and verify \cite{zheng2026shonan235,qian2025bridging}.

We refer to these structured representations collectively as symbolic world models. A symbolic world model explicitly represents entities, attributes, spatial and temporal relations, behavioural states, and environmental constraints in machine-interpretable forms such as scene graphs, semantic maps, knowledge graphs, object-centric representations, digital twins, or domain-specific symbolic abstractions \cite{xu2017scene}. Unlike latent neural embeddings, symbolic world models expose explicit semantics that can be queried, validated against specifications, shared across software modules, and refined as additional observations become available \cite{huang2025laser}. Importantly, symbolic world models establish the semantic interface between stochastic neural perception and formally analysable reasoning, enabling downstream testing, verification, planning, and runtime monitoring to operate over explicit semantic objects rather than opaque neural representations \cite{huang2026esca,qian2026humaninspiredneurosymbolicworldmodeling}.

Recent vision-language and multimodal foundation models substantially improve symbolic world-model construction by grounding perceptual observations into semantically meaningful concepts \cite{li2022grounded}. Beyond recognising individual objects, they infer attributes, affordances, interactions, intentions, and contextual knowledge from multimodal observations, reducing the manual engineering traditionally required for semantic perception pipelines \cite{driess2023palm,bravo2023open,minderer2022simple}. Rather than directly controlling the CPS, foundation models therefore function as semantic extraction and abstraction engines that continuously enrich structured world models.

Within the proposed Foundation Model Assurance Stack, symbolic world models become shared assurance artefacts across the system lifecycle. They provide the semantic foundation for behavioural specifications and domain-specific languages during system construction, support conformance relations, behavioural coverage, and structured scenario generation during testing, and expose formally analysable interfaces for contract verification, runtime monitor synthesis, and safety constraint synthesis. Consequently, testing, verification, and runtime assurance are unified around the same semantic representation rather than disconnected model-specific abstractions.

Constructing trustworthy symbolic world models remains challenging because perception uncertainty, sensor noise, partial observability, and distribution shifts inevitably propagate into symbolic abstractions \cite{qian2026humaninspiredneurosymbolicworldmodeling}. Furthermore, semantic grounding between physical observations and symbolic concepts is often incomplete in open-world environments, creating uncertainty that cannot be eliminated through perception alone. These challenges motivate explicit neuro-symbolic interfaces, uncertainty propagation mechanisms, and runtime assurance strategies that treat symbolic world models as assurance boundaries rather than unquestionable ground truth.

\subsection{Neural--Symbolic Interfaces and Contract Specifications}

Symbolic world models become useful assurance artefacts only when the boundary between neural and symbolic components is made explicit. A neural--symbolic interface therefore specifies far more than an output format. It defines the semantic meaning of exchanged information, including object vocabularies, coordinate systems, units, temporal scope, confidence or uncertainty representations, provenance, and the treatment of unknown or ambiguous observations. For example, a perception component should not merely output that an object is a pedestrian; it should expose the evidence supporting that classification, its spatial and temporal validity, an uncertainty estimate, and the operational conditions under which the prediction is expected to remain reliable. Such interface semantics prevent downstream reasoning, planning, and assurance components from treating uncertain neural predictions as unquestionable symbolic facts.

To make these interfaces machine interpretable and reusable throughout the assurance lifecycle, we advocate describing them using explicit contract specifications following the assume--guarantee paradigm~\cite{benveniste2012contracts}. Rather than constituting formal guarantees by themselves, these specifications document the assumptions under which a learning-enabled component is expected to operate and the properties that its outputs are intended to satisfy. Assumptions may include sensor quality, environmental conditions, latency bounds, distribution shift, or operational design domains, while expected properties describe admissible output domains, uncertainty representations, temporal consistency, calibration requirements, and response-time constraints. Since learning-enabled components are inherently stochastic, these specifications naturally accommodate probabilistic, set-valued, scenario-dependent, or uncertainty-aware descriptions rather than purely Boolean properties. Recent work on state-dependent conformal perception bounds illustrates how statistically derived uncertainty bounds can be incorporated into interface specifications connecting neural perception with symbolic reasoning~\cite{pmlr-v288-waite25a}.

Within the Foundation Model Assurance Stack, symbolic world models, interface specifications, and contract specifications become shared architectural artefacts that are consumed throughout the downstream assurance process. Structured world models provide the semantic basis for behavioural specifications and domain-specific languages, while interface and contract specifications expose explicit assumptions, abstractions, and constraints that can subsequently guide conformance-based testing, specification-guided scenario generation, interface verification, contract verification, runtime monitor synthesis, and safety constraint synthesis. Consequently, all assurance activities operate over a common set of architectural abstractions rather than constructing disconnected representations for testing, verification, and runtime assurance independently.

Explicit interfaces should also expose uncertainty and interface failures rather than silently propagating potentially incorrect symbolic abstractions. Whenever interface assumptions are violated, observations become unavailable, uncertainty exceeds acceptable bounds, or semantic inconsistencies arise, the component should explicitly report unknown or degraded-status outputs together with their associated confidence and provenance. Such architectural support allows downstream testing, verification, and runtime assurance mechanisms to reason systematically about degraded operation and uncertainty. Scenario-dependent probabilistic abstractions further capture how neural behaviour varies across environmental conditions and provide richer architectural descriptions for later system-level reasoning~\cite{watson2025scenario}. Likewise, explicit contract specifications establish the architectural foundation upon which contract composition and refinement can subsequently determine whether local component properties satisfy system-level requirements~\cite{benveniste2012contracts}.

Foundation models can substantially reduce the engineering effort required to construct these architectural artefacts. They can propose candidate abstractions, derive interface schemas, extract requirements from natural-language documentation, map domain knowledge to structured predicates, and identify missing assumptions or semantic inconsistencies. However, consistent with the architecture in Figure~\ref{fig:assurance-stack}, foundation models should be regarded as assistants for producing interface and contract specifications rather than trusted authorities for validating them. Generated specifications must therefore be checked for syntactic correctness, type consistency, satisfiability, traceability to system requirements, and conformance with observed system behaviour. Neural--symbolic interfaces and their contract specifications should consequently be treated as evolving, versioned assurance artefacts that co-evolve with the model, training data, operational domain, and accumulated runtime evidence.

\subsection{Symbolic Reasoning, Planning, and Control}
Structured world models provide an explicit interface for reasoning about
system behaviour, but trustworthy CPS must ultimately map such symbolic
abstractions back into physical actions. This creates a
\emph{symbolic--physical gap}: sensor observations must first be abstracted
into symbolic representations, decisions must be derived over these
representations, and the resulting decisions must then be translated into
physical control, with potential information loss at each boundary
\cite{zheng2026shonan235}. Neuro-symbolic architectures make these
interfaces explicit, allowing reasoning and decision making to operate over
semantically meaningful states rather than directly over opaque neural
representations \cite{zheng2025neurostrata}.

Foundation models can support this process without becoming authoritative
decision makers. They can assist in interpreting complex contexts and
constructing candidate symbolic artefacts, while symbolic reasoning
mechanisms and formal tools provide explicit constraints and independently
checkable semantics. This division of responsibility reflects a broader
principle for trustworthy CPS: foundation-model-supported reasoning should
not be conflated with formally guaranteed reasoning, and generated outputs
should be treated as constrained artefacts rather than trusted oracles
\cite{zheng2026shonan235}.

This separation naturally extends to planning and control. Hierarchical
neuro-symbolic architectures can connect symbolic mission-level planning,
semantic reasoning, and low-level perception and control while preserving
explicit interfaces between these layers \cite{zheng2025neurostrata}.
Such decomposition enables specifications and constraints to guide
decision making across abstraction levels and creates assurance boundaries
at which assumptions can be tested, verified, and monitored.

Concrete examples include neurosymbolic finite and pushdown automata, which
combine neural perception with explicit state and stack structures for multimodal
reasoning~\cite{pmlr-v288-sasaki25a}. Neuro-symbolic behavior trees (NSBTs) similarly
embed neural components in an explicit control structure and support verification
of safety and liveness properties formalized in temporal logic~\cite{pmlr-v288-serbinowska25a}. These architectures illustrate how explicit symbolic structure can expose intermediate
decisions and create targets for assurance.

Within the Foundation Model Assurance Stack, symbolic reasoning, planning,
and control connect structured world models to system execution, while
testing, verification, and runtime assurance provide complementary checks
on this process. Developing verified foundation-model-assisted planners
and scalable assurance across these neural--symbolic--physical interfaces
remains an important open research direction
\cite{zheng2026shonan235}.

\section{Testing and Validation of AI-Enabled Systems}
Testing constitutes the empirical pillar of the Foundation Model Assurance
Stack. A central outcome of Shonan Seminar No.~235 was that the challenge
for learning-enabled CPS is not simply to generate more tests, but to establish
whether observed system behaviour conforms to meaningful system-level
specifications and safety objectives~\cite{zheng2026shonan235}. Conventional
metrics based on input diversity, neural activation, or aggregate benchmark
performance provide limited evidence about whether a CPS satisfies its
operational requirements, particularly under rare interactions, distribution
shift, and failures that emerge across perception, reasoning, planning, and
control.

We therefore advocate \emph{conformance-driven validation} over the structured
assurance artefacts established in the architectural pillar. As illustrated in
Figure~\ref{fig:assurance-stack}, testing consumes symbolic world models,
specifications, interfaces, and contracts, and uses foundation models as
constrained assistants for specification mining, structured scenario generation,
and test synthesis. Symbolic and formal mechanisms then provide the
conformance relations, constraints, oracles, and adequacy criteria needed to
evaluate generated tests. Testing consequently produces more than pass/fail
outcomes: it identifies failure modes, counterexamples, violated assumptions,
behavioural coverage gaps, and empirically calibrated operating thresholds.
These artefacts flow in two directions across the assurance stack. Counterexamples
and constraint-validation results refine the verification mechanisms in Pillar~3,
while observed failure modes and calibrated thresholds inform runtime monitors,
safety envelopes, and switching conditions in the Simplex-based Runtime
Assurance Layer.

\subsection{Conformance-Based Testing over Structured World Models}

The foundation of the testing pillar is conformance-based testing. Rather than
evaluating a learning-enabled component solely through prediction accuracy or
benchmark performance, conformance testing asks whether the behaviour of the
integrated CPS satisfies explicit requirements, assumptions, and safety
properties. This distinction is particularly important because system-level
failures frequently arise not from an isolated incorrect prediction, but from
violations of operational assumptions, unsafe interactions among components,
or sequences of individually plausible decisions that collectively violate a
safety requirement.

Symbolic world models provide the semantic domain over which such conformance
relations can be defined. Objects, attributes, spatial and temporal relations,
system states, environmental conditions, and decision outcomes can be expressed
as structured predicates rather than raw sensor values. Testing can therefore
evaluate whether observed state transitions and execution traces conform to
mission requirements, architectural contracts, traffic rules, physical
constraints, or temporal safety properties. This directly addresses the
limitation identified at Shonan that neural coverage provides weak evidence for
system-level safety, while explicit symbolic world models make coverage, rule
checking, fault localisation, and conformance relations meaningful
~\cite{zheng2025neurostrata,zheng2026shonan235}.

This semantic representation also provides a structured test space. Instead of
perturbing pixels or sampling simulator parameters independently, test generation
can manipulate meaningful entities, relations, events, and environmental
conditions while preserving the constraints of the operational domain. For
autonomous driving, for example, scenarios can vary interactions among vehicles,
pedestrians, traffic infrastructure, road geometry, and environmental conditions;
for UAVs, they can vary terrain semantics, obstacles, visibility, and mission
constraints. Scenario generation thereby becomes an exploration of the same
symbolic state space used by reasoning and verification, rather than an
independent input-generation problem.

This alignment is important for compositional assurance. Architectural contracts
identify assumptions and admissible behaviours; verification exposes constraints
and properties whose boundaries require empirical validation; and testing
systematically explores the corresponding semantic regions. Conversely, observed
violations reveal where assumptions, abstractions, or contracts fail to match
physical behaviour. Conformance testing therefore provides the empirical bridge
between the symbolic abstractions of Pillar~1 and the formal guarantees sought
in Pillar~3.

\subsection{Foundation-Model-Assisted Specification and Test Synthesis}

A persistent bottleneck in testing learning-enabled CPS is the construction of
precise specifications and executable tests from heterogeneous assurance
knowledge. Safety requirements, operational procedures, regulations, accident
reports, and design assumptions are often expressed in natural language or other
unstructured forms, while testing requires machine-interpretable constraints,
oracles, and executable scenarios. Foundation models offer a powerful mechanism
for bridging this gap: they can extract candidate requirements, identify
behavioural constraints, interpret multimodal evidence, and synthesise candidate
test descriptions from sources that are difficult to process using conventional
program analysis alone \cite{deng2025target,luo2026safe, ma2025specgen}. However, consistent with the Foundation Model Assurance
Stack, these outputs should be treated as candidate assurance artefacts rather
than authoritative specifications or directly executable tests. Their correctness
must be established through independently checkable symbolic, formal, or
execution-based mechanisms.

This distinction motivates a constrained neuro-symbolic workflow for
foundation-model-assisted testing. Rather than asking a foundation model to
directly generate simulator scripts or other low-level test artefacts, we advocate
an intermediate representation based on domain-specific languages (DSLs),
symbolic world models, and structured specifications. Foundation models first
translate unstructured assurance knowledge into candidate structured artefacts;
these artefacts are then checked for syntactic validity, semantic consistency,
physical feasibility, and conformance with available domain constraints before
being instantiated into executable tests. The intermediate symbolic layer
therefore serves two purposes: it constrains the generative freedom of the
foundation model, while exposing explicit semantics that can be analysed,
verified, transformed, and traced back to their source requirements.

This generate--check--execute workflow should be iterative rather than one-shot.
Recent work in software testing has gone beyond one-shot test generation by
invoking foundation models when conventional test generation reaches a coverage
plateau~\cite{lemieux2023codamosa}. Generated test cases can subsequently be
executed and repaired by foundation models using feedback from conventional and
symbolic tools~\cite{schafer2023empirical}. These studies suggest a useful
division of responsibility for assurance-critical systems: foundation models
provide semantic interpretation and generative exploration, whereas symbolic
and conventional software-engineering tools provide constraints, oracles, and
measurable feedback. Failed validation or execution can then be returned to the
foundation model for repair, producing a closed
generate--check--execute--refine loop rather than relying on unconstrained
generation.

TARGET~\cite{deng2025target} illustrates this principle for
specification-driven testing of autonomous driving systems. Natural-language
traffic regulations are first interpreted and transformed into a compact DSL
representation rather than being translated directly into simulator scripts.
The resulting structured specifications are validated against the DSL and
subsequently instantiated into executable scenarios through symbolic route
search and parameter synthesis. This separation between
foundation-model-assisted interpretation and symbolically constrained execution
improves generation accuracy while preserving traceability from regulatory
requirements to concrete system tests. More importantly, it turns traffic rules
into explicit conformance objectives against which the behaviour of an
autonomous driving system can be evaluated.

SAFE~\cite{luo2026safe} extends the same principle from normative requirements
to empirical evidence of real-world failures. Instead of relying on free-form
scenario prompts, SAFE uses foundation models to interpret multimodal accident
reports, including textual descriptions and road sketches, and constructs
structured crash abstractions capturing participants, interactions,
environmental conditions, and causal relationships. These abstractions are
subsequently instantiated into executable scenarios that reproduce or vary
safety-critical behaviours observed in real incidents. Together, TARGET and
SAFE illustrate two complementary sources of test specifications:
\emph{prescriptive} knowledge describing how a system should behave and
\emph{descriptive} evidence revealing how systems and environments fail in
practice. Both are mediated through structured symbolic representations rather
than trusted foundation-model outputs.

The same workflow applies to specification mining. Foundation models can
propose candidate invariants, preconditions, postconditions, behavioural rules,
test oracles, and scenario constraints from documentation or observed
executions. Yet a plausible generated specification is not necessarily a correct
one. Recent work on analysing traditional software treats foundation-model
outputs as hypotheses rather than authoritative specifications
~\cite{ma2025specgen}. Candidate hypotheses are checked by verifiers and
systematically refined until a verifiable candidate is obtained. For
learning-enabled CPS, this principle is even more important because requirements
may be incomplete, context dependent, or valid only within particular
operational conditions. Generated specifications should consequently retain
their provenance, assumptions, and validation state, allowing human review,
testing, and formal analysis to establish progressively stronger assurance
evidence.

Foundation-model-assisted specification and test synthesis should therefore be
understood as an interaction across the assurance stack rather than an isolated
testing technique. The architectural pillar supplies symbolic world models,
interfaces, and domain abstractions that constrain what can be generated;
foundation models expand the candidate space of specifications and tests; and
verification supplies satisfiability checks, formal constraints, counterexamples,
and other independently checkable feedback. Executing the resulting tests, in
turn, reveals specification gaps, violated assumptions, and previously unknown
failure modes that can refine both architectural and verification artefacts.
These failure modes become the subject of behavioural coverage and adequacy
analysis in the next subsection and ultimately contribute to the empirical
calibration of runtime assurance boundaries. In this way, foundation models
provide generative breadth, while symbolic structure, verification, and execution
feedback provide the control necessary to turn generated artefacts into credible
assurance evidence.

\subsection{Behavioural Coverage, Failure Modes, and Test Adequacy}
Test generation alone provides little assurance unless accompanied by an
adequacy argument. Conventional software coverage criteria and their neural
analogues are insufficient when the relevant failures are defined at the level
of system behaviour. Prior studies have shown that increased neuron-activation
coverage does not necessarily improve fault detection
~\cite{harel2020neuron}, while surprise- and novelty-based criteria provide
alternative measures of unexplored model behaviour
~\cite{kim2019guiding,kim2023evaluating}. The Shonan discussions therefore
identified symbolic world-model coverage as a more semantically grounded
direction for learning-enabled CPS~\cite{zheng2026shonan235}.

Within the assurance stack, behavioural adequacy should be defined relative to
the system's specifications and hazards. At the semantic level, coverage can
measure whether relevant entities, attributes, relations, state transitions, and
interaction patterns in the symbolic world model have been exercised. At the
safety level, adequacy should capture whether testing has exercised known
failure modes, hazardous interactions, rule violations, and combinations of
conditions associated with unacceptable outcomes. At the assurance-boundary
level, testing should deliberately explore regions close to contract boundaries,
uncertainty limits, verified safety envelopes, and transitions between nominal
and degraded operation.

This perspective changes the role of a discovered failure. A collision, unsafe
landing, or violated temporal property is not merely counted as another failing
test; it should be associated with the symbolic state, environmental conditions,
interface assumptions, uncertainty levels, and execution history under which the
failure emerged. Clustering and characterising such observations yields explicit
\emph{failure modes} that can refine specifications and verification targets and,
critically, identify the conditions under which the high-performance controller
can no longer be trusted at runtime.

\subsection{Testing as Assurance Evidence: From Counterexamples to Runtime Thresholds}

Testing should ultimately produce assurance evidence rather than an isolated
collection of test results. Within the Foundation Model Assurance Stack, this
evidence connects all three pillars and the Runtime Assurance Layer.
Architectural abstractions determine what can be observed and tested, while
verification supplies contracts, invariants, safety constraints, and candidate
monitoring conditions. Testing then evaluates these artefacts against executable
system behaviour and the physical or simulated environment.

The interaction with verification is inherently bidirectional. Formal analysis
can identify boundary conditions, constraint surfaces, or assumptions that
require empirical exploration, thereby directing test generation toward
assurance-critical regions rather than uniformly sampling the behavioural space.
Conversely, testing produces counterexamples that expose incomplete
specifications, invalid assumptions, abstraction errors, or discrepancies between
formal models and implemented behaviour. These counterexamples can be fed back
to refine contracts, symbolic abstractions, verification conditions, and
subsequent test-generation strategies, forming a counterexample-guided assurance
loop.

Testing also provides a critical bridge from design-time assurance to runtime
operation through \emph{threshold calibration}. Runtime monitors frequently
depend on quantitative conditions---for example confidence, uncertainty,
distance to a safety boundary, timing margins, or deviations from expected
behaviour---for which the appropriate intervention threshold cannot be obtained
from a formal specification alone. Systematic testing around known failure modes
and verified safety boundaries can empirically characterise how these quantities
evolve as behaviour transitions from nominal to unsafe. The resulting calibrated
thresholds do not constitute formal guarantees; rather, they provide empirical
operating evidence that complements verified monitor logic and safety constraints
synthesised in Pillar~3.

This distinction is important for the Runtime Assurance Layer in
Figure~\ref{fig:assurance-stack}. Pillar~3 synthesises the monitor logic and
enforceable safety constraints, whereas testing characterises failure modes and
calibrates the operating thresholds under which those mechanisms should trigger.
At runtime, the Simplex architecture operationalises both forms of evidence:
the monitor evaluates the verified properties using calibrated observations, and
shielding or switching transfers control to the verified safe controller when
the applicable safety conditions are violated. Testing is therefore not merely
a pre-deployment activity; it provides empirical evidence that determines how
design-time assurance artefacts are instantiated and exercised during runtime.

\section{Verification}
\label{SecVerification}
This section develops the verification pillar of the Foundation Model Assurance Stack. We first identify tractable verification targets, then consider compositional verification across contracts, interfaces, and transformations, foundation-model-assisted verified synthesis, and the synthesis of runtime monitors and safety constraints for operational use.

\subsection{Tractable Verification Targets}
A central challenge for foundation-model-enabled CPS is deciding \emph{what}
should be formally verified. Conventional verification reasons about potentially
infinite or intractably large spaces of behaviours
~\cite{DBLP:books/daglib/0020348,DBLP:books/daglib/0019162}, whereas testing
provides empirical evidence over a finite set of executions
~\cite[Section~5]{DBLP:books/daglib/0040158}. Directly extending this paradigm
to an entire foundation model is problematic: foundation models are large,
stochastic, and semantically open-ended, while the properties relevant to CPS
safety often concern interactions among perception, world modelling, reasoning,
planning, and physical execution rather than the behaviour of an isolated
model~\cite{DBLP:journals/corr/abs-2512-19007}. The verification objective
should therefore not be to prove the foundation model correct in the abstract,
but to identify assurance-critical parts of the system for which meaningful
properties can be stated and independently checked.

The Shonan discussions suggest organising such targets according to
\emph{formal tractability} \cite{zheng2026shonan235}. Rather than treating ``verification of foundation
models'' as a monolithic problem, verification should prioritise
components and transformations whose correctness admits explicit
specifications and verification oracles. These include symbolic world models,
interface-level guardrails, contracts, representation transformations, formal
reasoning components, and generated software or specifications for which
correctness or equivalence can be independently checked. This principle does
not exclude verification of neural or foundation models themselves; rather, it
focuses verification effort where formal reasoning can establish properties
that contribute directly to a system-level assurance argument.

Existing neural network verification methods provide concrete component-level
targets. NNV and NNV 2.0 support reachability-based verification of neural
networks and learning-enabled CPS~\cite{tran2020nnv,lopez2023nnv2}. ImageStars and
relaxed reachability extend such analysis to convolutional perception and semantic
segmentation~\cite{tran2020imagestars,tran2021segmentation}. NNV3 expands the
supported architectures and application domains~\cite{tumlin2026nnv3}.
Complementary methods such as $\beta$-CROWN combine bound propagation with
branch-and-bound search for neural-network robustness
verification~\cite{wang2021betacrown}. More broadly, existing results illustrate scalability to realistic networks, particularly those of the scale that arise in CPS for control and perception tasks~\cite{DBLP:journals/corr/abs-2512-19007}. These results establish properties under
specified input sets and model assumptions, which must be related to the
overall CPS.

Closed-loop verification also connects neural outputs to plant dynamics. Controllers trained using reinforcement learning have been verified with star set reachability to
verify safety properties in learning-enabled CPS, including an Advanced Emergency Braking
System (AEBS) case study~\cite{tran2019emsoft}. This provides a concrete precedent
for stating safety guarantees over a neural controller composed with a model of
the plant. The broader community has verified many other neural networks control systems (NNCS) in ARCH-COMP~\cite{sasaki2026arch}.

Neuro-symbolic architectures make this strategy practical by exposing explicit
assurance boundaries around otherwise opaque learned components. Symbolic
world models provide formally analysable representations of the environment;
contracts make component assumptions and guarantees explicit; interfaces
specify what information may cross component boundaries; and transformation
layers expose how neural outputs are converted into representations consumed
by symbolic reasoning. These artefacts turn parts of an end-to-end learning
pipeline into explicit verification obligations. 
Consequently, assurance does not require complete verification of the foundation model itself. Instead, the system can constrain and verify how the model's outputs are represented, interpreted, and used in safety-critical decisions.

The resulting verification strategy is therefore selective and compositional.
Properties should be verified at the abstraction level where they can be stated
precisely and where their satisfaction has identifiable consequences for system
safety. Verification may establish local properties of a component, but these
properties become useful for system assurance only when their assumptions,
interfaces, and transformations are also justified. This motivates the
contract-, interface-, and transformation-centred verification developed next.
It also establishes the scope for foundation-model-assisted verified synthesis:
foundation models may propose specifications, transformations, or executable
artefacts, but correctness must ultimately rest on independently checkable
verification evidence rather than on the model that generated them.

\subsection{Compositional Verification of Contracts, Interfaces, and Transformations}
Once tractable verification targets have been identified, the central challenge
is to compose local guarantees into meaningful system-level assurance. In a
neuro-symbolic CPS, correctness does not depend solely on whether individual
components satisfy their local properties, but also on whether the assumptions
under which those properties hold remain valid as information flows across
perception, world modelling, reasoning, planning, and control. We therefore
treat contracts, interfaces, and transformations as complementary verification
targets that collectively define the assurance boundaries of the architecture.

Contracts make these dependencies explicit by associating components with
assumptions about their inputs and environments and guarantees about their
outputs and behaviour. For example, a perception component may provide
bounded-error guarantees on its estimates
~\cite{DBLP:journals/pacmpl/AstorgaHMM23}, while reasoning or planning
components may guarantee satisfaction of symbolic invariants, temporal
properties, or operational constraints under corresponding assumptions.
This naturally supports assume--guarantee reasoning
~\cite{DBLP:conf/nato/Pnueli84}: components can be verified locally, while
system-level reasoning establishes whether the guarantees provided by one
component satisfy the assumptions required by another. Such decomposition is
particularly important for learning-enabled CPS, for which constructing and
verifying a faithful monolithic model of the complete system is generally
intractable.

Recent data-driven methods analyse probabilistic robustness of semantic
segmentation networks~\cite{hashemi2025scaling} and learn reach-set geometry to
tighten probabilistic neural-network verification~\cite{sasaki2026learning}. Such
results can inform quantitative component contracts when their input-distribution,
model-fidelity, and statistical assumptions are explicit, where their probabilistic
guarantees must retain those assumptions during system-level composition.

Contracts alone, however, are insufficient if the interfaces through which
components exchange information do not preserve the semantics on which those
contracts depend. Neuro-symbolic architectures introduce explicit interfaces
between neural perception and symbolic reasoning, as well as among world
modelling, planning, and control. Verification at these boundaries should
establish not only structural compatibility---for example, schema and type
consistency---but also semantic obligations such as the preservation of
uncertainty, spatial and temporal relationships, provenance, and assumptions
required by downstream components. Neuro-symbolic verification provides one
mechanism for expressing properties that span neural and symbolic components
by embedding neural networks within logical specifications
~\cite{DBLP:conf/ijcai/XieKN22}. This makes it possible to state properties
relating neural outputs to the symbolic representations consumed downstream,
rather than restricting verification to properties of an isolated neural
network.

A particularly critical class of interface obligations concerns
\emph{transformations} between representations. The physical--symbolic boundary
is not a passive communication channel: perception outputs may be abstracted
into predicates, object relations, scene graphs, or other symbolic world-model
elements before formal reasoning begins. An error introduced during this
transformation can satisfy the syntax of an interface while nevertheless
invalidate subsequent reasoning. In NeuroSymLand, for example, perception
outputs are transformed into probabilistic predicates and structured semantic
scene representations before being consumed by symbolic reasoning
~\cite{qian2026humaninspiredneurosymbolicworldmodeling}. Transformation
verification should therefore establish appropriate preservation properties
between source and target representations, including consistency of object and
relation mappings, uncertainty information, and safety-relevant spatial or
temporal semantics.

The same principle extends beyond neural-to-symbolic abstraction. Assurance
artefacts may undergo multiple transformations across the stack: requirements
may be formalised into specifications or DSLs, symbolic states may be translated
between reasoning and planning representations, and verified properties may
later be compiled into runtime monitors or enforceable safety constraints.
Each transformation introduces a proof obligation concerning what information
or semantics must be preserved. Making these obligations explicit is essential
because downstream verification is only meaningful under the assumptions
established by the preceding transformation.

Contract, interface, and transformation verification consequently form a
compositional assurance chain. Local component guarantees are meaningful only
if their environmental assumptions are discharged; interface guarantees are
meaningful only if exchanged representations retain the required semantics; and
downstream reasoning guarantees are meaningful only if the transformations
constructing their symbolic inputs preserve the properties on which those
guarantees depend. This decomposition also supports incremental assurance:
when a learned component is updated, the complete CPS need not necessarily be
re-verified if the new component can be shown to satisfy the same externally
visible contracts, interface obligations, and transformation properties.

Testing and verification interact directly at these boundaries. Conformance
testing can expose violated assumptions, inconsistent world-model
constructions, unexpected interface behaviours, and abstraction errors that
invalidate existing verification obligations. Such counterexamples can refine
contracts, interface specifications, and transformation properties. Conversely,
verified assumptions, boundary conditions, and invariants provide semantically
meaningful objectives for subsequent testing. Contracts, interfaces, and
transformations therefore serve as shared assurance artefacts through which
empirical evidence from Pillar~2 and formal reasoning in Pillar~3 mutually
refine one another.

\subsection{Foundation-Model-Assisted Verified Synthesis}
Foundation models can complement formal verification by generating candidate
artefacts whose correctness is subsequently established by conventional
verification mechanisms. This is particularly attractive for assurance tasks
that remain labour intensive despite having machine-checkable correctness
criteria, including specification and invariant generation, code and model
translation, auto-formalisation, and synthesis of executable artefacts from
formal requirements. The key distinction is that foundation models expand the
space of candidate solutions but do not establish their correctness. Consistent
with the Shonan discussions, foundation-model assistance should therefore be
used primarily where generated outputs can be checked against explicit
specifications, semantic equivalence relations, or other independently
executable verification oracles \cite{zheng2026shonan235}.

We advocate a \emph{propose--verify--refine} workflow for such tasks. A
foundation model first proposes a candidate program, specification, model,
proof artefact, or transformation. An independent verification mechanism then
checks the relevant correctness obligation using, for example, type checking,
SMT solving, model checking, symbolic execution, equivalence checking, or a
proof assistant. Successful verification produces machine-checkable evidence
for the generated artefact; failure produces counterexamples, failed proof
obligations, type errors, or other symbolic feedback that can guide subsequent
generation. Iterating this process turns foundation-model generation from an
unconstrained source of plausible artefacts into a search procedure operating
within a formally checked solution space.

This separation between generation and verification is especially important for
translation and synthesis. A foundation model may produce a program that is
syntactically valid and functionally plausible while subtly changing the
semantics of its source, or generate a formal specification that is internally
consistent but fails to capture the intended requirement. Translation
validation provides a useful precedent: rather than proving a complex
translator correct once and for all, each generated output can be checked
against its source~\cite{pnueli1998translation,necula2000translation}. More
generally, proof-carrying approaches demonstrate how an artefact produced by
an untrusted generator can be accepted on the basis of independently
checkable evidence~\cite{necula1997pcc}. Foundation models can therefore
remain outside the trusted computing base provided that accepted outputs are
validated by sufficiently small and trustworthy checking mechanisms.

Learned semantic representations can improve the efficiency of this workflow
without replacing formal verification. For example, a foundation model may
associate a natural-language requirement with previously verified formal
specifications that express similar requirements, or identify correspondences
between source and target programs intended to preserve the same behaviour.
These associations can help retrieve relevant verified examples and guide the
generation of candidate artefacts that are more likely to satisfy the required
correctness conditions. They do not, however, establish semantic equivalence.
The generated artefact must still be checked against its source requirement or
program using an appropriate formal verification mechanism.
For example, \textsc{ProofBridge}
~\cite{jana2025proofbridge} combines semantically aligned representations with
retrieval and verifier-guided repair for auto-formalisation. Crucially, semantic
alignment is used to improve generation rather than to certify it: formal
checking remains responsible for determining whether the resulting artefact is
acceptable.

This perspective leads to a broader notion of \emph{certificate-producing AI}.
Where a task admits machine-checkable correctness conditions, the desirable
output of foundation-model-assisted synthesis is not merely an apparently
correct artefact, but an artefact accompanied by evidence that can be checked
independently of the model that generated it. Depending on the task, such
evidence may take the form of an equivalence certificate, proof object,
satisfied verification conditions, or another checkable witness. This
distinguishes foundation-model-supported verification from claims that the
foundation model itself has been formally verified: trust is placed in the
specification and checker, not in the stochastic generation process.

Within the Foundation Model Assurance Stack, verified synthesis consequently
connects foundation-model assistance with the formal guarantees of the
Verification pillar. It can transform requirements into checked formal
specifications, translate or synthesise system artefacts under explicit
correctness obligations, and produce verified artefacts that become inputs to
subsequent assurance activities. In particular, the same principle applies when
verified properties are transformed into operational assurance mechanisms:
runtime monitors and enforceable safety constraints may be synthesised with
foundation-model assistance, but their correctness must be established
independently before they are supplied to the Runtime Assurance Layer. The
synthesis of these two classes of runtime artefacts is considered next.

\subsection{Runtime Monitor and Safety Constraint Synthesis}

Formal verification is necessarily performed with respect to models,
specifications, and assumptions established at design time, whereas
learning-enabled CPS operate in environments whose states, uncertainties, and
interactions cannot be completely anticipated before deployment. The
Verification pillar therefore has a second role beyond establishing design-time
properties: it must translate properties that remain relevant during operation
into assurance artefacts that can be evaluated or enforced at runtime. In the
Foundation Model Assurance Stack, we distinguish two such outputs:
\emph{runtime monitors}, which make selected verification obligations
operationally observable, and \emph{safety constraints}, which define the
conditions within which system behaviour is permitted to remain.

Runtime monitor synthesis transforms formally specified properties into
executable observers over system executions. Runtime verification provides a
well-established foundation for this process, allowing execution traces to be
evaluated against formally specified properties
~\cite{DBLP:journals/jlp/LeuckerS09}. Within a neuro-symbolic CPS, monitorable
properties may include contract assumptions, safety invariants, temporal
requirements, uncertainty bounds, operational-domain conditions, and
consistency requirements over symbolic world-model states. Because these
properties are expressed over the explicit representations and interfaces
introduced in Pillar~1, monitor synthesis can target semantically meaningful
system states rather than opaque neural activations. A monitor can therefore
retain traceability to the specification, contract, or verification obligation
from which it was derived.

Not every verified property, however, is directly monitorable at runtime.
Some properties depend on latent physical states, future behaviour, or
information unavailable during execution, while others are too expensive to
evaluate within the timing constraints of a CPS. Monitor synthesis must
therefore preserve the semantics of the original verification obligation while
identifying an executable observation model: which symbolic variables and
events are observable, what history must be retained, how uncertainty is
represented, and under what conditions a monitor can soundly detect a
violation. This creates an important connection with interface and
transformation verification: the validity of a runtime monitor depends on the
correctness of the representations through which physical and neural states
become observable symbolic events.

Safety constraint synthesis complements runtime monitoring by deriving, from verified system properties, the conditions under which the learning-enabled system can operate safely.
Depending on the CPS and the available
formal models, these constraints may encode admissible states and actions,
temporal or spatial separation requirements, control-invariant conditions,
resource and timing bounds, or conservative conditions derived from verified
system properties. Their purpose is not to prescribe the nominal behaviour of
the high-performance learning-enabled controller, but to expose a formally
grounded boundary beyond which its behaviour should no longer be accepted
without intervention. Runtime safety envelopes around symbolic abstractions
are particularly attractive when the underlying perception or learning
components cannot themselves be completely verified, because they concentrate
formal reasoning on the conditions required for safe system operation.

Monitor and safety-constraint synthesis also provide an important point of
interaction between verification and testing. Verification determines the
logical structure of the properties and safety conditions that must be observed
or enforced, whereas testing provides counterexamples, observed failure modes,
and empirical evidence near these boundaries. In particular, the threshold
calibration produced by Pillar~2 can parameterise monitor conditions or
intervention criteria whose numerical values cannot be derived from the formal
model alone. Such calibration does not turn empirical thresholds into formal
guarantees, nor should it replace formally established safety constraints.
Rather, it connects verified properties with evidence about how the implemented
system behaves as it approaches the corresponding assurance boundaries.

Foundation models may assist this synthesis process by translating requirements,
contracts, counterexamples, and verification results into candidate monitoring
rules or safety constraints. However, the same principle developed in the
previous subsection applies: generated artefacts are proposals rather than
assurance evidence. Candidate monitors must be checked against their source
properties, and candidate safety constraints must be validated using appropriate
formal analysis before they become part of the assurance architecture.
Foundation-model assistance can therefore reduce the engineering effort of
constructing operational assurance artefacts without placing the model itself
inside the trusted assurance boundary.

The outputs of the Verification pillar are consequently explicit:
\emph{verified properties}, \emph{runtime monitors}, and \emph{safety
constraints}. Importantly, their synthesis does not itself constitute runtime
assurance. Pillar~3 establishes and prepares these artefacts, while the
Simplex-based Runtime Assurance Layer in Section~5 operationalises them during
system execution. There, monitors observe the deployed system and the safety
constraints provide the basis for shielding or switching to a verified safe
controller when intervention is required. This separation maintains a clear
boundary between the design-time construction of formally justified assurance
mechanisms and their runtime execution under uncertainty.

\section{Runtime Assurance through the Simplex Architecture}
\label{SecRuntimeAssurance}
Design-time architecture, testing, and verification provide complementary
assurance evidence, but none alone can guarantee that a learning-enabled CPS
will encounter only behaviours anticipated before deployment. Open
environments, distribution shift, perception uncertainty, incomplete models,
and interactions among learning and physical components inevitably leave a
residual assurance gap. The Shonan discussions therefore identified runtime
enforcement and verified safety envelopes as practical complements to
design-time assurance, particularly when complete verification of realistic
learning-enabled pipelines remains out of reach \cite{zheng2026shonan235}. We use the Simplex
architecture~\cite{Sha2001,BakRTAS2009} as the operational foundation for
closing this gap; representative extensions and instantiations are discussed
in~\cite{10.1145/3457335.3461712,RamakrishnaHBKD,PhanNFM2020}.

As illustrated in Figure~\ref{fig:assurance-stack}, the Runtime Assurance Layer
is distinct from the three assurance pillars. Rather than producing additional
design-time evidence, it operationalises their outputs during system execution.
The architectural pillar supplies the system structure, neural--symbolic
interfaces, and symbolic world models through which runtime state is exposed;
testing supplies empirically observed failure modes and calibrated thresholds;
and verification supplies verified properties, runtime monitors, and safety
constraints. These artefacts converge in a Simplex-style closed loop comprising
a high-performance learning-enabled controller, a runtime monitor, a verified
safe controller, and shielding or switching mechanisms. The purpose of this
layer is therefore not to make the learning-enabled controller itself fully
verifiable, but to constrain its operational authority so that high-performance
behaviour is permitted only while the assumptions and safety conditions
supporting that behaviour continue to hold.

\subsection{Simplex-Based Operational Assurance}
At the centre of runtime operation is the \emph{learning-enabled controller},
which represents the high-performance decision-making path constructed through
the neuro-symbolic architecture of Section~\ref{secFoundationModels}. The term
controller is used broadly here: it may comprise foundation-model-assisted
perception, symbolic world-model construction, reasoning, planning, and
lower-level control rather than a single neural component. For example, in
NeuroSymLand, neural perception is transformed into probabilistic symbolic
representations over which safety reasoning is performed before landing
decisions influence UAV control \cite{qian2026humaninspiredneurosymbolicworldmodeling}. Such pipelines provide adaptability and
semantic reasoning capabilities that would be difficult to reproduce with a
small verified controller, but their dependence on stochastic perception,
learned abstractions, and open-world assumptions means that they should not be
given unconstrained authority over safety-critical behaviour.

The \emph{runtime monitor} determines whether that authority can continue to
be exercised. It instantiates the monitor logic synthesised in
Section~\ref{SecVerification} over the observable system state, including
symbolic world-model elements, interface states, uncertainty estimates, and
execution histories. Rather than treating monitoring as an independent anomaly
detector, we view it as the runtime counterpart of the assurance obligations
developed across the stack. It evaluates whether relevant contract assumptions
remain valid, whether verified safety conditions continue to hold, and whether
the system is approaching empirically characterised failure regions. This
connection is particularly important for neuro-symbolic CPS because structured
representations allow violations to be associated with explicit entities,
relations, events, assumptions, and uncertainty rather than only with opaque
neural activations.

When continued operation of the learning-enabled controller can no longer be
justified, assurance relies on a \emph{verified safe controller}. In contrast
to the high-performance path, the safe controller deliberately sacrifices
capability and optimality for analysability and predictable behaviour. Depending
on the CPS, it may implement emergency braking, collision avoidance, a
safe-hover or abort mode, restricted-speed operation, or another conservative
control policy whose safety properties can be established over a defined
operating envelope. This asymmetry is fundamental to Simplex: assurance does
not require the advanced controller and safe controller to provide equivalent
functionality. It requires the safe controller to provide a sufficiently simple
and verified means of maintaining or recovering safety when reliance on the
advanced controller is no longer justified.

The connection between these controllers is provided by \emph{shielding and
switching}. Switching follows the classical Simplex principle
~\cite{Sha2001,BakRTAS2009}: when the monitored state indicates that continued
operation of the learning-enabled controller would violate the applicable
safety conditions, control authority is transferred to the verified safe
controller. Shielding instead constrains the advanced controller without
necessarily replacing it, for example by rejecting or modifying actions that
would leave an admissible safe region. These mechanisms may also be combined:
local violations may be handled by restricting individual actions, whereas
more severe assumption failures, uncertainty, or proximity to a safety boundary
may require transfer to the safe controller. Representative Simplex-based
runtime-assurance mechanisms and extensions illustrate how such architectures
can preserve high performance during nominal operation while maintaining a
formally justified fallback path
~\cite{10.1145/3457335.3461712,RamakrishnaHBKD,PhanNFM2020}.

Real-time reachability provides a concrete basis for Simplex switching
decisions by analysing future plant behaviour within the runtime computation
budget~\cite{bak2014rtss,johnson2016tecs}. Action-level intervention is also
illustrated by shields that enforce temporal-logic safety specifications during
reinforcement learning~\cite{DBLP:conf/aaai/AlshiekhBEKNT18}. RusTReach connects real-time reachability with neurosymbolic reinforcement
learning for safe autonomous navigation~\cite{pmlr-v288-potteiger25a}. It
illustrates how reachability-based runtime assurance can be integrated with a
learned navigation policy and symbolic planning.

The resulting runtime loop separates \emph{capability} from \emph{authority}.
The learning-enabled controller proposes high-performance behaviour, but its
authority to affect the physical system remains conditional on the runtime
assurance mechanism. The monitor evaluates the conditions under which that
authority is valid, while shielding or switching prevents unacceptable actions
or transfers authority to the safe controller when those conditions cease to
hold. This separation is particularly suitable for foundation-model-enabled
CPS: it permits powerful but incompletely verified models to contribute to
system behaviour without making their correctness a prerequisite for every
system-level safety claim.

\subsection{Integrating Assurance Artefacts at Runtime}
The distinguishing role of the Runtime Assurance Layer in the Foundation Model
Assurance Stack is that its decisions are grounded in assurance artefacts
accumulated across the preceding pillars rather than in a standalone confidence
score or anomaly detector. Architecture, testing, and verification contribute
different forms of evidence, and these forms should remain distinguishable
because they support different parts of the runtime decision.

First, the architectural pillar determines \emph{what can be observed and
interpreted}. Symbolic world models expose objects, relations, system states,
and environmental conditions; neural--symbolic interfaces specify their
semantics, uncertainty, provenance, coordinate systems, and validity
conditions; and contracts identify the assumptions under which individual
components are intended to operate. These artefacts provide the observation
model on which runtime assurance depends. A monitor can only determine that a
contract assumption or safety condition has been violated if the relevant
physical and neural state has been exposed through an appropriate runtime
interface. The architectural pillar therefore supplies the \emph{system
structure and interfaces} through which operational safety properties become
observable.

Second, testing supplies empirical knowledge about \emph{where and under what
conditions the implemented system fails}. Conformance testing and structured
scenario exploration identify failure modes, violated assumptions, and
behavioural regions in which the learning-enabled controller becomes
unreliable. Testing around these regions can additionally calibrate quantitative
conditions such as uncertainty, distance to a safety boundary, timing margins,
or other indicators associated with transitions from nominal to unsafe
behaviour. These calibrated thresholds provide operational evidence for
triggering runtime intervention. Importantly, they remain empirical evidence:
a threshold estimated from testing does not itself constitute a formal safety
guarantee.

Third, verification determines \emph{what properties must continue to hold}.
As developed in Section~\ref{SecVerification}, contract, interface, and
transformation verification establish formally justified properties, while
runtime monitor synthesis and safety constraint synthesis translate applicable
properties into operational artefacts. Runtime monitors specify what conditions
should be observed during execution, and safety constraints define the
admissible boundary within which the learning-enabled controller may retain
authority. Verification therefore provides the formal structure of runtime
assurance, while testing helps instantiate and calibrate that structure against
the behaviour of the implemented system.

These inputs converge at runtime. The monitor evaluates verified properties
over the structured state exposed by the architecture, using empirical
calibration where appropriate to interpret uncertainty and proximity to known
failure regions. The resulting monitor state determines whether the proposed
behaviour of the learning-enabled controller remains admissible under the
applicable safety constraints. If it does, high-performance operation
continues. If a proposed action violates an enforceable constraint, a shield
may restrict or modify that action; if the assumptions supporting the
learning-enabled controller cease to hold or safety can no longer be maintained
through local intervention, switching transfers authority to the verified safe
controller. Runtime assurance thus closes the loop from design-time evidence
to physical system behaviour.

Uncertainty is integral to this process rather than an additional runtime
mechanism. Perception error, partial observability, distribution shift, and
stochastic model behaviour mean that the system may lose confidence in the
assumptions supporting a safety argument before an explicit safety violation
occurs. Uncertainty represented through the neural--symbolic interfaces can
therefore form part of monitored contract assumptions and intervention
conditions. Increasing uncertainty may tighten admissible operating conditions,
cause individual actions to be shielded, or ultimately trigger transfer to the
safe controller. The purpose is not to treat uncertainty estimates themselves
as safety guarantees, but to prevent the system from continuing to rely on a
learning-enabled component after the evidence supporting its assumed competence
has become insufficient.

The output of the Runtime Assurance Layer is consequently not another
specification, test result, or verification certificate, but \emph{constrained
system behaviour}: actions reaching the physical CPS have passed through the
operational assurance boundary established by the Simplex architecture. During
nominal operation, this preserves the capability of the learning-enabled
controller; as the system approaches or crosses its justified operating
boundary, shielding or switching progressively restricts that authority in
favour of verified safe behaviour. Runtime assurance thereby closes the
Foundation Model Assurance Stack by turning architectural structure, empirical
evidence, and formal guarantees into continuous operational control over how
learning-enabled behaviour is allowed to affect the physical world.

\section{Research Agenda: From Foundation Models to Certifiable AI Systems}
\label{sec:agenda}
The Foundation Model Assurance Stack provides a systems-oriented framework for
integrating neuro-symbolic architecture, testing, verification, and runtime
assurance. Its broader implication, however, is that trustworthy
foundation-model-enabled CPS cannot be achieved by independently improving
individual assurance techniques. The central research problem is to establish
an assurance chain from human and regulatory intent, through neural perception
and symbolic abstraction, to formal analysis and ultimately physical system
behaviour. A failure at any boundary can invalidate guarantees established
elsewhere in the stack.

This observation suggests a shift in research emphasis. The objective should
not be to make every foundation model intrinsically ``safe'' or fully verified
before it can participate in a safety-critical system. For general-purpose
models, such an objective remains technically remote and, for many open-ended
tasks, difficult even to specify. Instead, research should determine
\emph{where formal structure can be introduced, what properties can be
independently checked, how local evidence composes into system-level assurance,
and how the remaining uncertainty can be bounded at runtime}. Foundation
models can then contribute capability within explicit assurance boundaries,
while symbolic representations, formal analysis, empirical testing, and
runtime enforcement determine the authority that those capabilities receive.

Building on the Foundation Model Assurance Stack and the research priorities
emerging from the Shonan discussions \cite{zheng2026shonan235}, we identify four interconnected
challenges. They follow the lifecycle of an assurance argument: defining what
the system should do, establishing trustworthy abstraction boundaries,
composing evidence into system-level guarantees, and maintaining and
certifying those guarantees throughout deployment.

\subsection{Challenge 1: From Informal Intent to Executable Assurance Specifications}
The first challenge is the persistent \emph{specification bottleneck}. Formal
verification presupposes properties to verify, conformance testing requires
criteria against which behaviour can be judged, and runtime assurance requires
conditions that can be monitored and enforced. Yet safety requirements for
learning-enabled CPS rarely arrive as complete formal specifications. They are
distributed across regulations, engineering requirements, safety cases,
operational procedures, incident reports, domain expertise, and implicit human
expectations. Statements such as ``maintain a safe distance'' or ``select a
safe landing site'' acquire precise meaning only when contextual factors,
uncertainty, operational modes, and environmental assumptions are made
explicit.

Foundation models create an opportunity to address this bottleneck because
they can connect heterogeneous natural-language and multimodal knowledge with
structured representations. However, specification generation should not be
treated as unconstrained text-to-logic translation. A generated formula can be
syntactically valid while misrepresenting the source requirement, omitting an
important exception, or introducing assumptions unsupported by the physical
system. The research problem is therefore to develop \emph{assurance-aware
specification engineering}: foundation models propose candidate requirements,
contracts, predicates, scenarios, and DSL artefacts, while domain constraints,
formal consistency checks, empirical evidence, and human review progressively
establish their validity.

Several deeper questions follow. How should specifications represent
uncertainty and context rather than only Boolean requirements? How should
conflicting requirements extracted from regulations, expert knowledge, and
operational data be detected and resolved? How can requirements be traced from
their natural-language origin to contracts, test obligations, verification
conditions, and runtime monitors? And how should specifications evolve when
testing or deployment reveals previously unknown operating conditions?

The long-term objective is a \emph{living assurance specification}: a
traceable, machine-processable collection of contracts, constraints,
environment assumptions, and safety properties shared across the lifecycle.
A requirement mined from a regulation should be traceable to the tests that
exercise it, the verification obligations that formalise it, and the runtime
conditions that enforce it. 
Specification engineering would thereby provide a common foundation for testing, verification, and runtime assurance, rather than being performed independently for each assurance activity.

\subsection{Challenge 2: Closing the Neural--Symbolic--Physical Assurance Gap}
The second challenge concerns the assurance boundaries on which the entire
stack depends. Neuro-symbolic decomposition makes learning-enabled CPS more
analysable by converting neural outputs into explicit objects, relations,
events, predicates, and world states. Yet introducing a symbolic abstraction
does not by itself make the abstraction correct. If a pedestrian is omitted
from a scene graph, an uncertain relation is converted into a deterministic
predicate, or a symbolic action is incorrectly mapped into physical
coordinates, downstream reasoning may be formally correct with respect to a
world model that is wrong with respect to the physical world.

This exposes a fundamental \emph{neural--symbolic--physical assurance gap}.
Assurance must address not only individual components but also the semantic
fidelity of the transformations between physical observations, learned
representations, symbolic abstractions, decisions, and physical actions.
These boundaries are especially important for multimodal foundation models,
where information from vision, language, maps, sensors, and prior knowledge
may be fused through representations whose semantics are difficult to
characterise formally.

A major research direction is therefore to develop \emph{assurance-carrying
interfaces}. Rather than transmitting only predictions or symbolic values,
interfaces should expose the assumptions, uncertainty, provenance, validity
region, and transformation history needed by downstream assurance mechanisms.
A symbolic predicate should, where relevant, carry evidence about how it was
grounded; a world-model relation should expose the uncertainty under which it
was inferred; and a transformation should make explicit the properties it is
required to preserve. This would allow testing, verification, and runtime
monitoring to reason not only about the symbolic state but also about the
conditions under which that state can legitimately be trusted.

This agenda raises several fundamental problems. We need theories for
property-preserving neural-to-symbolic abstraction, compositional uncertainty
propagation, multisensor consistency, abstraction refinement, and semantic
equivalence across heterogeneous representations. Empirical conformance
testing will remain necessary where physical grounding cannot be proved
formally, while formal verification can establish structural and logical
properties once assumptions have been made explicit. The important research
question is therefore not whether neural or symbolic reasoning alone should
be trusted, but how evidence crossing their boundary can retain sufficient
semantics for subsequent assurance claims to remain valid.

Success here would turn symbolic world models from useful intermediate
representations into explicit \emph{assurance boundaries}. They would provide
a common semantic substrate on which behavioural coverage can be defined,
component contracts composed, runtime conditions monitored, and failures
localised. This is a prerequisite for scaling the remainder of the Foundation
Model Assurance Stack beyond isolated demonstrations.

\subsection{Challenge 3: From Component Evidence to Compositional System Assurance}
The third challenge is to determine how heterogeneous assurance evidence can
support meaningful system-level claims. Current research frequently evaluates
individual components in isolation: perception robustness is measured against
perturbations, testing reports empirical failure rates or coverage, formal
verification proves properties of restricted models, and runtime monitoring
detects violations during execution. Each form of evidence is useful, but none
automatically implies that the complete CPS is safe. The central scientific problem is therefore how to combine assurance evidence from individual components into defensible system-level assurance claims.

The Foundation Model Assurance Stack suggests that compositional assurance
should be organised around explicit contracts and symbolic interfaces.
Component-level evidence must establish assumptions and guarantees that are
meaningful to neighbouring components and ultimately relevant to system-level
safety properties. This requires stronger theories for assume--guarantee
reasoning under probabilistic perception, uncertainty propagation across
interfaces, abstraction refinement, and modular re-assurance when components
or foundation models are updated.

Composition must also extend across assurance methods. Testing and verification
should not remain parallel activities. Formal contracts and invariants can
identify semantically meaningful test obligations and boundary conditions,
while testing can expose counterexamples, violated assumptions, and operating conditions that reveal where formal models need to be refined. Similarly,
verification and runtime assurance should be connected through monitor and
safety-constraint synthesis, while deployment evidence should reveal when
assumptions used during verification no longer hold. The resulting process is
a closed assurance loop rather than a sequence of disconnected development
stages.

Foundation models can accelerate this loop, but their role requires a precise
trust boundary. They can propose specifications, invariants, abstractions,
translations, proof candidates, and counterexample explanations, but acceptance
must depend on an independent oracle whenever correctness is formally
expressible. This motivates a general \emph{propose--check--counterexample--
refine} architecture for Foundation model-assisted assurance. Parsers, simulators, theorem
provers, SMT solvers, model checkers, static analyses, and differential tests
provide increasingly strong forms of feedback depending on the task. In
certificate-producing workflows, an foundation model may generate a candidate artefact, but
the certificate and its independent checker---not the foundation model---establish the
accepted guarantee. Existing work coupling semantically aligned models with
formal verification harnesses illustrates the potential of this direction
~\cite{jana2025proofbridge}.

An important research objective is therefore to develop a \emph{compositional framework for assurance evidence}: a principled account of what different evidence means, under which
assumptions it remains valid, and how multiple pieces of evidence combine into
a defensible system claim. Formal proof, statistical evidence, calibrated
uncertainty, simulation results, field observations, and runtime intervention
records provide fundamentally different strengths of evidence and should not
be collapsed into a single confidence score. Instead, their dependencies and
limitations should be explicit. Such a framework would also enable incremental
assurance: when a foundation model or subsystem is replaced, only the affected
contracts and evidence dependencies should need to be re-established rather
than requiring complete recertification.

\subsection{Challenge 4: From Assurance Evidence to Lifecycle Certification and Operational Trust}
The fourth challenge is translating technical assurance results into evidence
that remains meaningful to operators, regulators, and certification
authorities. This is particularly difficult for learning-enabled CPS because
the object being assured is not necessarily static. Foundation models may be
updated, operating environments change, data distributions drift, and
previously unseen interactions may invalidate assumptions established during
development. Certification based solely on a pre-deployment snapshot is
therefore poorly matched to the lifecycle of these systems.

Future assurance frameworks should connect each system-level safety claim to
the evidence supporting it and to the assumptions under which that evidence
remains valid. Architecture can identify component and interface boundaries;
testing can provide empirical evidence about known failure modes and operating
regions; verification can establish formal properties and generate monitors
and safety constraints; and runtime assurance can record whether those
assumptions continue to hold during operation. The resulting assurance case
should expose these dependencies rather than presenting safety as a single
binary property.

This perspective motivates \emph{continuous assurance cases} in which
deployment evidence can strengthen, challenge, or invalidate existing claims.
A distribution shift may not immediately imply unsafe operation, for example,
but it may invalidate the empirical calibration supporting a runtime threshold.
A component update may preserve its interface contract and therefore require
only local re-assurance, whereas a changed abstraction boundary may invalidate
several downstream claims. Developing principled rules for such evidence
evolution is essential if certification is to scale to systems that change
after deployment.

Human oversight must be incorporated into the same framework. Operators need
evidence that supports appropriate trust calibration: why an intervention was
triggered, which assumption failed, what uncertainty increased, and whether
control has been transferred to a safe controller. Conversely, human
intervention and operational experience become additional evidence about
system competence boundaries. Research is therefore needed on predictive
handoff, interpretable process-level evidence, assurance-case evolution,
auditability, tool qualification, and mappings between formal artefacts and
standards such as those governing safety-critical automotive, robotic, and
industrial systems.

The long-term goal is not merely a system that passes a certification test,
but a system whose \emph{assurance state} can be maintained, inspected, and
re-established throughout its operational lifecycle.

\subsection{A Staged Path toward Certifiable Foundation-Model-Enabled Systems}
These challenges suggest that progress toward certifiable AI should be staged
according to \emph{assurance tractability}, rather than model capability. The
appropriate question is not ``how large a foundation model can we verify?''
but ``how far can we expand the boundary within which correctness is explicitly
specified, independently checked, and operationally enforced?'' This principle
provides a consistent way to organise both near-term research and longer-term
grand challenges.

In the near term, research should concentrate on assurance tasks with \textit{explicit
correctness oracles}. These include specification and annotation generation for
structured domains, syntactic and semantic consistency checking, symbolic
world-model coverage, verification of small guardrail components, invariant
proposal, and certificate-producing translation of code and formal artefacts.
Foundation models can already provide substantial value in these settings
because generated candidates can be checked independently. Thus, the immediate
objective is not fully autonomous assurance, but reliable
\emph{Foundation Model-assisted assurance} in which generation is separated from acceptance.

The next stage is \emph{compositional system assurance}. Here, the principal
challenge shifts from checking isolated artefacts to establishing how
contracts, symbolic world models, transformations, test evidence, and runtime
constraints compose across realistic CPS pipelines. Research should demonstrate
that assurance survives component replacement, environmental variation, and
distribution shift within explicitly stated limits. Benchmarks should
therefore move beyond isolated model accuracy or neural robustness and measure
end-to-end assurance properties: specification traceability, semantic
coverage, contract satisfaction, counterexample discovery, uncertainty
propagation, monitor effectiveness, and preservation of safety under runtime
intervention.

Existing competitions provide useful starting points for this evaluation
agenda. The International Verification of Neural Networks Competition (VNN-COMP) compares neural-network verification tools on shared
benchmarks~\cite{DBLP:journals/corr/abs-2512-19007}, while the hybrid systems verification competition (ARCH-COMP) category on AI and NNCS (AINNCS)
evaluates mostly reachability analysis for neural network-controlled continuous
and hybrid plants~\cite{sasaki2026arch}. Extending such evaluations to the
contracts, semantic transformations, and runtime interventions of
foundation-model-enabled CPS remains a research objective. In both verification tasks, remaining challenges exist around specification, scalability, and generality of models that can be verified~\cite{johnson2025allerton}.

A subsequent stage is \emph{certifiable closed-loop autonomy}. At this stage,
design-time and runtime evidence become parts of a unified lifecycle assurance
case. Verified properties define the safety envelope; testing characterises
failure modes and calibrates uncertain boundaries; runtime monitors determine
whether assumptions remain valid; and Simplex-style shielding and switching
prevent high-performance learning-enabled controllers from exercising authority
outside their justified envelope. Certification then concerns not only whether
a component passed a test before deployment, but whether the complete
assurance mechanism continues to maintain its stated claims during operation.

Only after these foundations mature should the assurance boundary progressively
expand toward more ambitious direct verification of foundation-model
components. Small guardrail models and formally structured code or mathematical
tasks provide realistic starting points, followed by domain-specific planners,
scientific foundation models constrained by physical invariants, and
multimodal models operating within explicitly specified domains. Verification
of unrestricted general-purpose foundation-model behaviour remains a valuable
grand challenge, but it should not be a prerequisite for trustworthy deployment
of systems whose safety can instead be established compositionally.

This staged agenda changes the meaning of progress in trustworthy AI-enabled
CPS. Progress is not measured solely by stronger models, larger benchmarks, or
more sophisticated individual assurance tools. It is measured by the
progressive expansion of a \emph{machine-checkable assurance boundary}: from
individual specifications and interfaces, to neuro-symbolic pipelines, to
closed-loop autonomous systems, and ultimately to certifiable
foundation-model-enabled CPS. The Foundation Model Assurance Stack provides
the structure for this progression. Architecture establishes explicit
assurance boundaries; testing challenges them empirically; verification
establishes properties over the portions that are formally tractable; and
runtime assurance constrains physical behaviour when design-time guarantees
reach their limits. Foundation models increase capability throughout this
process, but independently checkable evidence determines trust.

\section{Conclusion}
Foundation models substantially expand the capabilities of learning-enabled
cyber-physical systems, but capability alone does not establish trust. Their
stochastic and opaque behaviour, together with uncertainty introduced across
perception, reasoning, planning, and physical execution, challenges
model-centric assurance and motivates a shift toward system-level engineering.
In this paper, we introduced the \emph{Foundation Model Assurance Stack} as a
systems-oriented framework for regaining engineering control over such systems.
Its central principle is to establish explicit assurance boundaries around
learning-enabled components through symbolic world models, neural--symbolic
interfaces, contracts, and structured representations, and to use these shared
artefacts to connect conformance-based testing, compositional verification,
and Simplex-based runtime assurance.

This perspective also defines a deliberate role for foundation models within
safety-critical CPS. Foundation models can provide semantic interpretation,
generative exploration, reasoning support, and synthesis capabilities, but
they should remain constrained assistants rather than trusted oracles.
Testing challenges their surrounding assumptions empirically; verification
establishes properties where correctness can be formally specified and
independently checked; and runtime assurance limits the authority of
learning-enabled behaviour when those assumptions or safety conditions cease
to hold. Trust therefore derives not from the apparent capability or confidence
of a foundation model, but from the independently checkable evidence and
operational constraints that govern how its outputs can affect the physical
system.

Looking forward, we argue that progress toward certifiable AI should be
measured by the progressive expansion of this \emph{machine-checkable assurance
boundary}: from executable specifications and trustworthy neural--symbolic
interfaces, through compositional system assurance, to closed-loop runtime
protection and lifecycle certification. Full verification of unrestricted
foundation models need not be a prerequisite for trustworthy deployment.
Instead, a practical path is to verify what is formally tractable, test what
must be established empirically, monitor what can change during operation, and
constrain behaviour when available evidence no longer justifies continued
authority. The resulting research agenda moves beyond isolated advances in
models, testing, or verification toward an integrated discipline of
\emph{Certifiable AI Engineering}, in which foundation models increase system
capability while independently checkable assurance evidence determines trust.



\bibliographystyle{ACM-Reference-Format}
\bibliography{acmart,ttj}

\appendix

\end{document}